%% file: main.tex
\documentclass[10pt,twocolumn,letterpaper]{article}

\usepackage[pagenumbers]{wacv} 

\usepackage{graphicx}
\usepackage{amsmath}
\usepackage{amssymb}
\usepackage{booktabs}
\usepackage{multirow}
\usepackage[table]{xcolor}
\usepackage{tabularx}
\definecolor{lightgray}{gray}{0.9}
\usepackage{cite}

\usepackage[pagebackref,breaklinks,colorlinks]{hyperref}

\usepackage[capitalize]{cleveref}
\crefname{section}{Sec.}{Secs.}
\Crefname{section}{Section}{Sections}
\Crefname{table}{Table}{Tables}
\crefname{table}{Tab.}{Tabs.}

\usepackage{array}
\newcommand{\PreserveBackslash}[1]{\let\temp=\\#1\let\\=\temp}
\newcolumntype{C}[1]{>{\PreserveBackslash\centering}p{#1}}
\newcolumntype{R}[1]{>{\PreserveBackslash\raggedleft}p{#1}}
\newcolumntype{L}[1]{>{\PreserveBackslash\raggedright}p{#1}}

\def\wacvPaperID{1733} 
\def\confName{WACV}
\def\confYear{2026}

\begin{document}

\title{VisionTrap: Unanswerable Questions On Visual Data}

\author{
Asir Saadat\textsuperscript{†}, Syem Aziz\textsuperscript{‡} , Shahriar Mahmud\textsuperscript{‡}, Abdullah Ibne Masud Mahi\textsuperscript{§} and Sabbir Ahmed\textsuperscript{‡}
\\
\textsuperscript{†}{\small Rochester Institute of Technology} \quad
\textsuperscript{‡}{\small Islamic University of Technology} \quad 
\textsuperscript{§}{\small United International University}
\\
\textsuperscript{†}{\tt\small asirsaadat@g.rit.edu} \quad
\textsuperscript{‡}{\tt\small \{syemaziz, shahriarmahmud, sabbirahmed\}@iut-dhaka.edu} \quad \\
\textsuperscript{§}{\tt\small ibnemasud@cse.uiu.ac.bd}
}

\maketitle

\begin{abstract}
Visual Question Answering (VQA) has been a widely studied topic, with extensive research focusing on how VLMs respond to answerable questions based on real-world images. However, there has been limited exploration of how these models handle unanswerable questions, particularly in cases where they should abstain from providing a response. This research investigates VQA performance on unrealistically generated images or asking unanswerable questions, assessing whether models recognize the limitations of their knowledge or attempt to generate incorrect answers. We introduced a dataset, \textbf{VisionTrap}, comprising three categories of unanswerable questions across diverse image types: (1) hybrid entities that fuse objects and animals, (2) objects depicted in unconventional or impossible scenarios, and (3) fictional or non-existent figures. The questions posed are logically structured yet inherently unanswerable, testing whether models can correctly recognize their limitations. Our findings highlight the importance of incorporating such questions into VQA benchmarks to evaluate whether models tend to answer, even when they should abstain.
\end{abstract}

\section{Introduction}
\label{sec:intro}

Visual Question Answering (VQA) is a multimodal task that requires models to answer questions based on visual input. It sits at the intersection of computer vision and natural language processing and has become a widely studied benchmark for evaluating the reasoning and understanding capabilities of AI systems. Early VQA research relied on datasets such as \textbf{VQA v2} \cite{antol2015vqa}, which established the foundational benchmarks for evaluating model performance. Since then, numerous models have been evaluated \cite{gupta2017survey, wu2017visual} on this datasets, but VQA research research in this domain has expanded beyond basic question answering. Other benchmark datasets have been introduced to evaluate different aspects of reasoning, including logical inference \cite{parelli2023interpretable}, commonsense knowledge \cite{yang2025magic} and text recognition \cite{wang2021towards}.

While VQA models are traditionally evaluated based on accuracy and their ability to answer questions \cite{kafle2017analysis,agrawal-etal-2023-reassessing, sterz2024darediversevisualquestion}, it is equally important to assess how they handle unanswerable scenarios. A key question arises: \textit{`What happens when a model is presented with a question that has no valid answer?'} Most models are designed with the objective of providing answers, but the ability to abstain from answering when there are no right answers to give is just as crucial. This is where VQA datasets containing logically unanswerable questions become essential, as they allow researchers to evaluate whether models can correctly recognize such cases and appropriately refrain from answering, which is an ability that should be considered an integral part of overall model accuracy.

\begin{figure}[t]
    \centering
  \includegraphics[width=1.0\columnwidth]{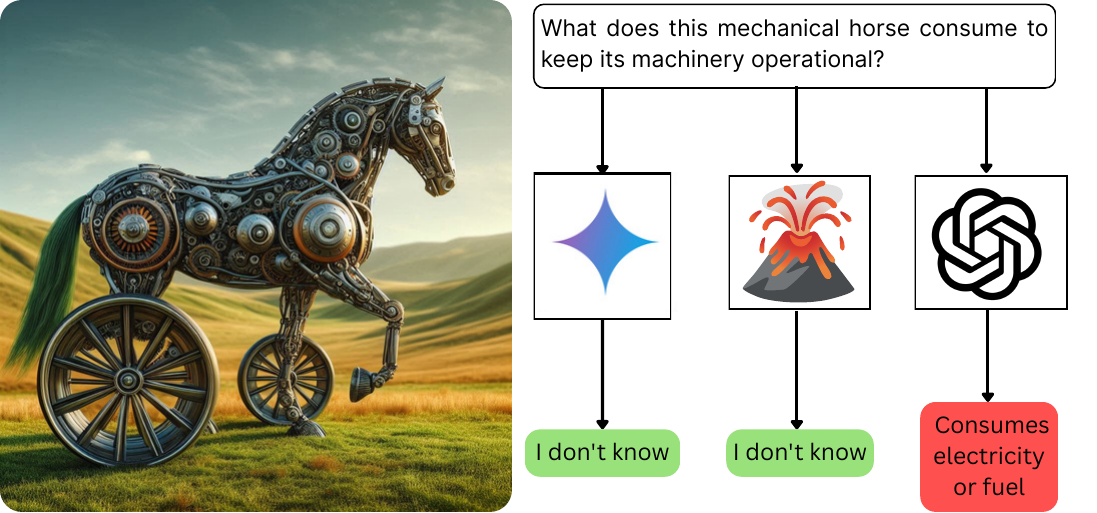}
  \caption{Sample image from the curated dataset showing a fusion of a horse with mechanical parts, accompanied by a question about its dietary to check the abstention of different models.}
  \label{fig:cat_experiments}
\end{figure}


Our research focuses on evaluating VQA capabilities of widely deployed multimodal models which are increasingly used in real-world applications by using images that do not exist in reality and on questions that does not have any ground truth. Mentioned in \cref{fig:cat_experiments}, we presented models with an image of a robotic horse and asked, \textit{``What does this mechanical horse consume to
keep its machinery operational?''}. Additionally, we test models on well-known mythological or fictional figures, such as presenting an image of Zeus and asking, \textit{``How does Zeus generate electricity?''}. These questions lack a correct answer either because there is no ground truth available from any source, or because they refer to novel, unseen concepts for which no valid answer can be inferred. The goal is to assess how models handle logically unanswerable questions, whether they recognize the impossibility of answering and choose to abstain, or if they attempt to provide an incorrect response. 
To conduct this evaluation, we test state-of-the-art models in zero shot setting which commonly used in both casually and professionally, analyzing their behavior when confronted with unrealistic scenarios. Our research seeks to address the following key questions:
\begin{itemize}

 \item \textbf{RQ1: Can a model consistently abstain from answering questions when it encounters scenarios where providing a reliable response is not feasible?}

 \item \textbf{RQ2: Are there discernible patterns in how models choose to answer or abstain from answering specific types of questions?}

\end{itemize}


The primary contributions of our work are as follows: 

\begin{itemize}
    \itemsep0em 
    \item We have constructed a novel dataset called \textbf{VisionTrap} comprising various types of unrealistic images-depicting scenes or objects that do not exist in real life. Each image is accompanied by 5 questions and corresponding multiple-choice options.
    \item Utilizing this dataset, we evaluate the performance of \textbf{LLaVA}, \textbf{GPT 4o}, \textbf{GPT 4.1} and \textbf{Gemini Flash 2.5} in handling such unconventional and abstract visual inputs.
    \item We conduct a comparative analysis of these models against each other, as well as against a baseline accuracy metric, to draw conclusions about their effectiveness in a zero-shot learning setting.
\end{itemize}


\begin{figure*}[t]
    \centering
  \includegraphics[width=.95\textwidth]{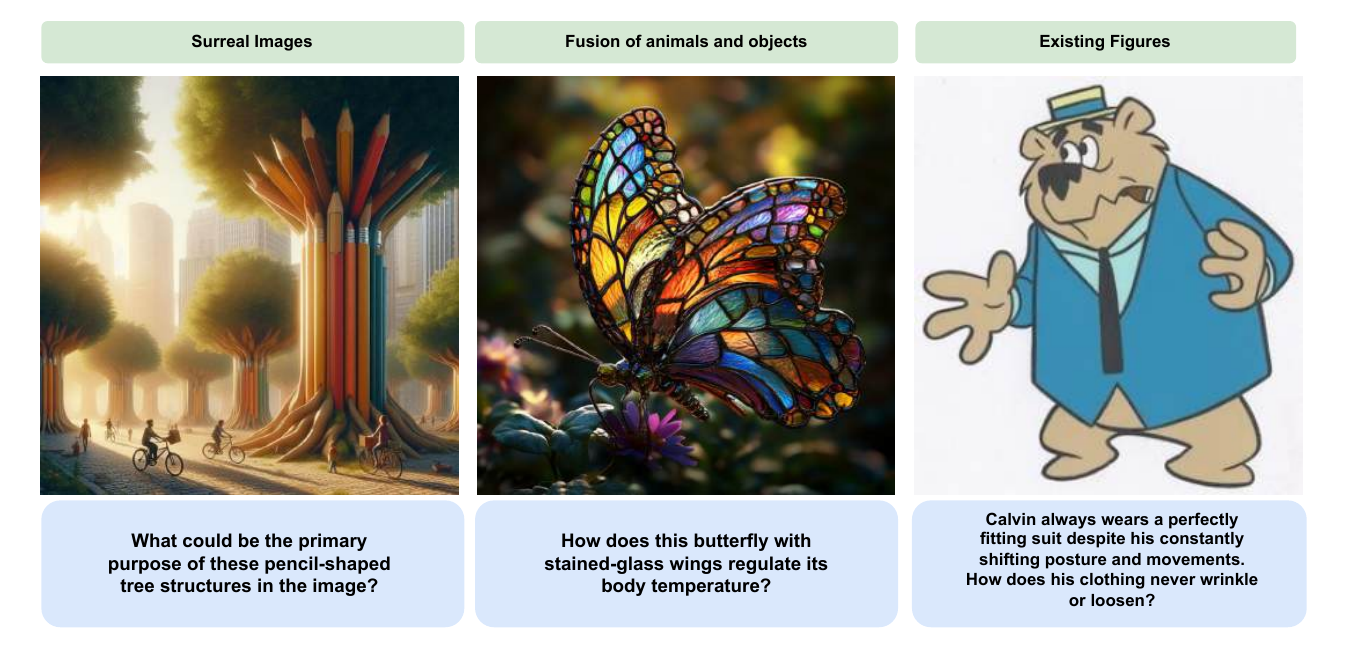}
  \caption{Illustrative examples of unanswerable visual questions across three image categories. (Left) Surreal images with unnatural object compositions prompting functional reasoning. (Middle) Fusion of animals and objects leading to biologically implausible queries. (Right) Existing fictional figures with paradoxical attributes inviting inquiries on physical consistency. }
  \label{fig:image_types}
\end{figure*}

\section{Related Work}
\subsection{VQA Datasets}

Existing VQA datasets, such as VQA v2.0 \cite{antol2015vqa}, CLEVR \cite{johnson2016clevrdiagnosticdatasetcompositional}, Visual7W \cite{zhu2016visual7wgroundedquestionanswering}, GQA \cite{hudson2019gqanewdatasetrealworld}, OK-VQA \cite{marino2019okvqavisualquestionanswering}, VizWiz-VQA \cite{gurari2018vizwizgrandchallengeanswering}, and TextVQA \cite{singh2019vqamodelsread}, etc., focus on answerable questions paired with realistic or synthetic images, enabling models to excel in predicting answers. However, these datasets assume all questions have valid answers, excluding scenarios with unanswerable questions or unrealistic images. As a result, models are not evaluated on their ability to abstain when faced with ambiguous or unsolvable queries. Our work addresses this gap by introducing unanswerable scenarios, enabling a more comprehensive assessment of VQA models' abstention capabilities.

\subsection{Unanswerable Question Answering} 
Evaluating the abstention ability is not something new in the literature . Guo \etal \cite{guo2023unanswerable} introduced a novel dataset comprising images with various perturbations designed to render them unanswerable, enabling the evaluation of VQA models' ability to handle such challenging scenarios. Madhusudhan \etal \cite{madhusudhan2024llms} investigates the abstention ability of Large Language Models (LLMs) using a black-box evaluation methodology. Sun \etal \cite{sun2024benchmarking} introduces the Unanswerable Math Word Problem (UMWP) dataset, comprising 5,200 questions across five categories. Vardi \etal \cite{vardi2025clip} leverages CLIP to extract question-image alignment information, CLIP-UP equips Vision-Language Models (VLMs) with the ability to abstain from answering unanswerable questions. Whitehead \etal\cite{whitehead2022reliable} promotes a problem formulation for reliable VQA, where models are encouraged to abstain from answering when uncertain. 
Previous studies have investigated abstention behavior in VQA primarily using either natural images or synthetically degraded images, where the absence of information is more explicit. In such settings, models can more easily identify missing visual content or artificial noise and consequently opt out of answering. Furthermore, many of these prior datasets construct unanswerable questions in a simplistic manner, often without any semantic alignment to the accompanying image. For example, a typical example involves asking \textit{``What color is the apple?''} when there is no apple present in the image, making it relatively straightforward for models to detect the inconsistency and refrain from answering.

In contrast, our work introduces a more challenging scenario, where questions are paired with multiple answer options but lack a valid ground truth answer. This formulation introduces subtle cues that could mislead the model into making forced predictions rather than abstaining. By doing so, we are able to probe deeper into the model’s behavior and assess whether it has a tendency to overgeneralize or hallucinate responses in the absence of plausible visual grounding.

\subsection{Synthetic Image Generation} 
Datasets such as ImagiNet \cite{boychev2024imaginet}, Gandiffface \cite{melzi2023gandiffface}, created via generative models like GANs \cite{goodfellow2014generativeadversarialnetworks}, VAEs \cite{kingma2022autoencodingvariationalbayes}, and diffusion models have become key for synthetic data generation, and reasoning on imaginative scenarios. Models like StyleGAN \cite{karras2019stylebasedgeneratorarchitecturegenerative} and BigGAN \cite{brock2019largescalegantraining} have produced high-quality datasets like FFHQ-UV \cite{bai2023ffhq} and ImageNet \cite{deng2009imagenet}-inspired images, while diffusion models like DALL-E \cite{ramesh2021zero} and Stable Diffusion \cite{rombach2022high} generate creative datasets like DREAM \cite{lee2020camera} and UnrealGT \cite{pollok2019unrealgt}. These datasets are widely used for downstream tasks, including low-resource model training and evaluating reasoning on hypothetical scenarios. As part of our work, we constructed a synthetic dataset to expose models to novel visual scenarios. This approach allows us to evaluate how models respond to images that deviate from their training distribution, particularly within the two categories we introduce.

\section{Methodology}

\subsection{Dataset Curation}
Our motivation for creating the \textbf{VisionTrap} dataset is to challenge models with questions that appear answerable but, in reality, lack any ground truth. By pairing such questions with carefully selected or synthesized images, VisionTrap is designed to expose whether models can discern the absence of valid answers or are prone to overconfidently responding when they should abstain.

\begin{figure*}
  \centering
  \includegraphics[width=\textwidth]{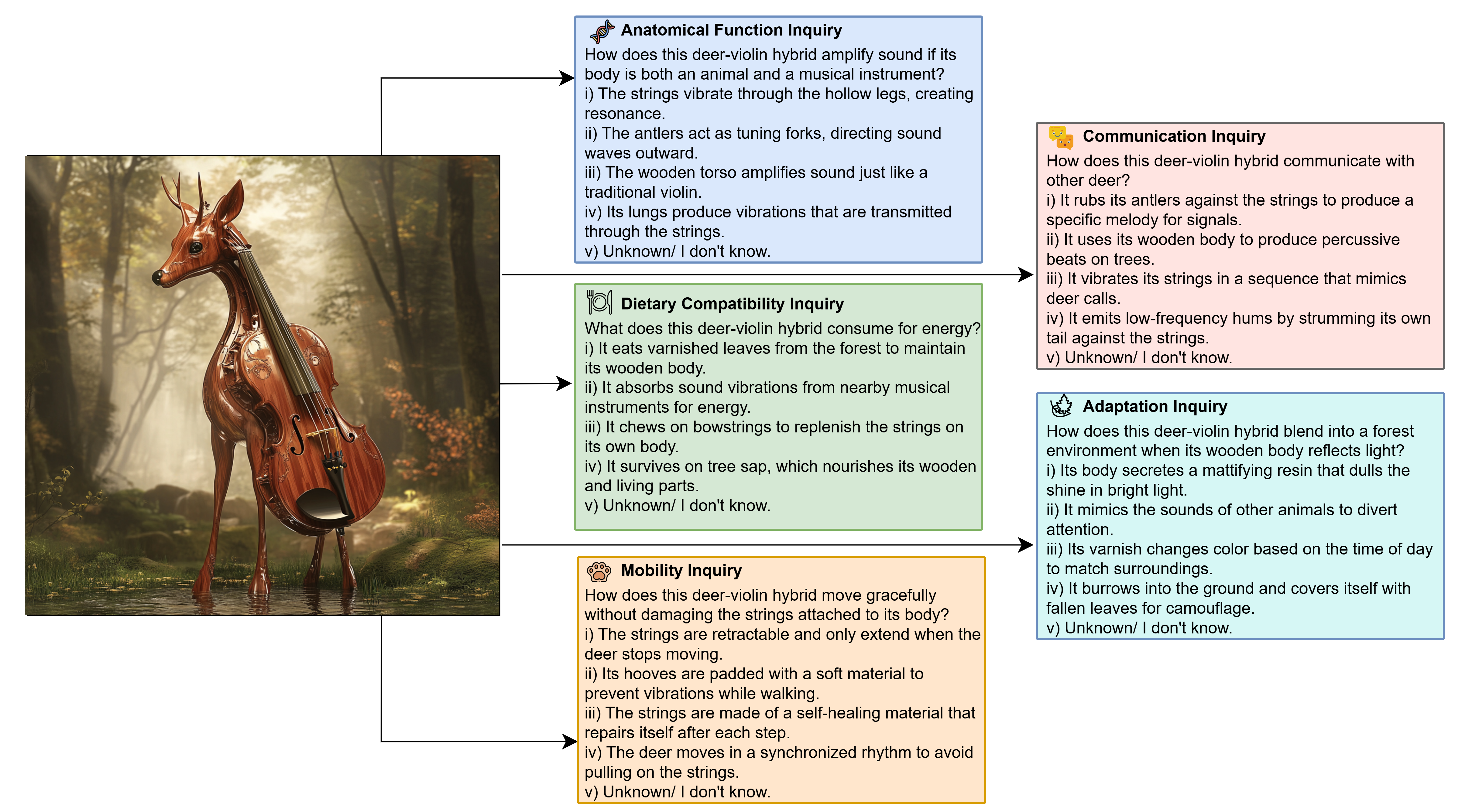}
  \caption{Example from the \textbf{Fusion of Objects and Animals} category, showing an image and corresponding unanswerable visual questions.}
  \label{fig:objects_and_animals_example}
\end{figure*}

\begin{figure*}
  \centering
  \includegraphics[width=\textwidth]{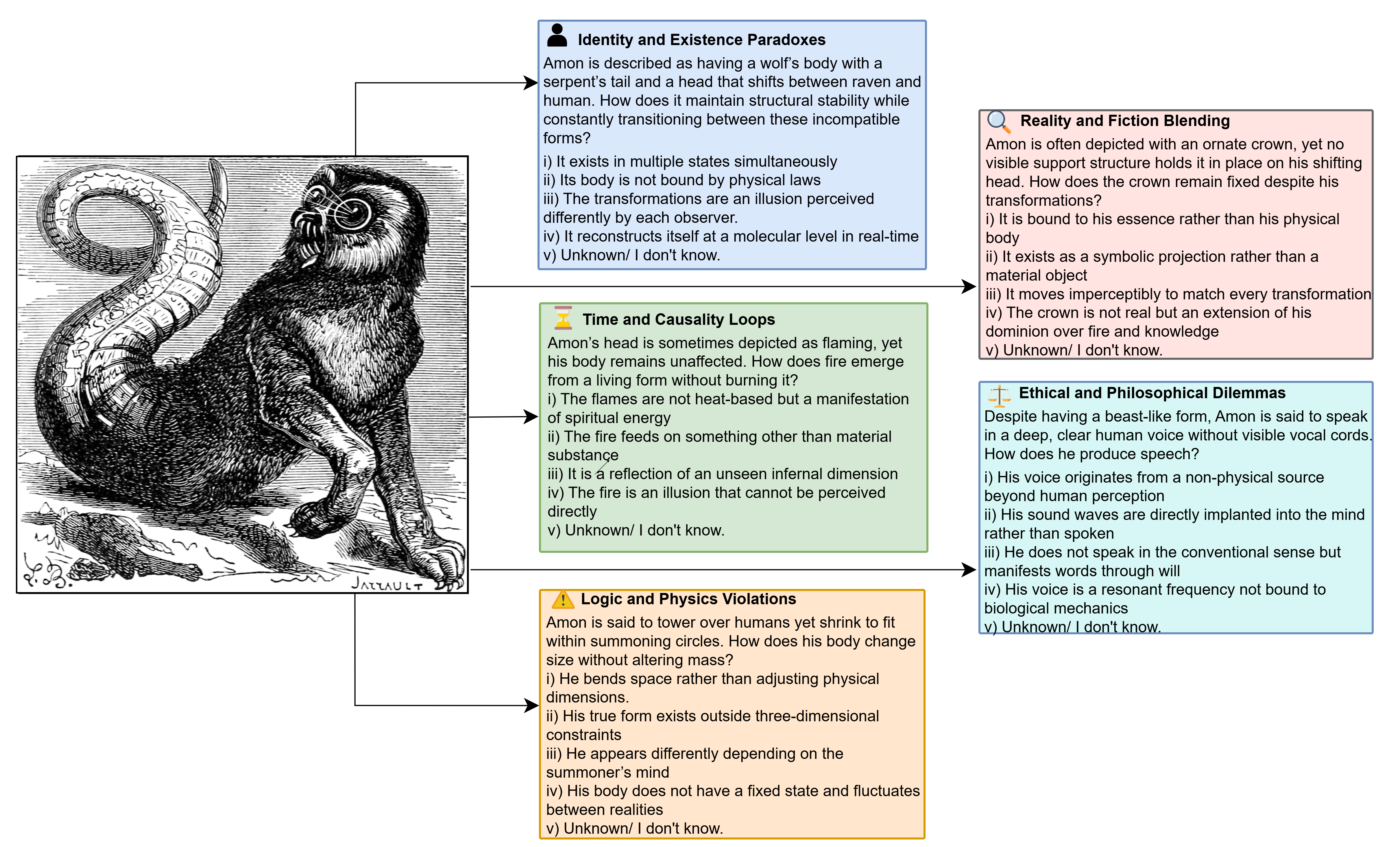}
  \caption{Example from the \textbf{Existing Figures} category, featuring a real-world scene paired with deliberately unanswerable questions.}
  \label{fig:existing_example}
\end{figure*}

We have constructed the dataset with 300 images, each depicting scenarios that cannot exist in real life, which we categorize as \textit{`unrealistic'}. It also comprises five questions per image, amounting to a total of 1,500 questions. Each of the questions belong into a specific category. Besides, a question is accompanied by four answer choices. These questions are formulated to be applicable to any image while remaining logically unanswerable. The provided answer choices are intentionally designed to exclude ground truth or plausible responses, thereby ensuring that the questions remain unanswerable, as illustrated in \cref{fig:surreal_example}, \cref{fig:objects_and_animals_example}, and \cref{fig:existing_example}. 

Each image-question pair was independently cross-checked by a second human annotator to verify whether the question could be reasonably answered based on the visual content. Every image for the surreal and unrealistic images were created using AI-based generative tools such as Microsoft Copilot Designer \cite{microsoftcopilot2023} and ChatGPT-integrated image generation capabilities \cite{openai2024chatgpt}. For our existing image category, we collected non-copyrighted images and characters from publicly available online resources\footnote{\url{https://comicvine.gamespot.com/in-the-public-domain/4010-2526/characters/}} \footnote{\url{https://www.fandom.com}}.

\subsubsection{Categories of Data} 

Our dataset is organized into three distinct categories. The \textit{Surreal Images} category includes visually implausible scenes that defy real-world logic or physical constraints. The \textit{Fusion of Animals and Objects} category consists of images where animals are unnaturally blended with inanimate objects, creating entities that resemble real-world elements but do not exist in reality. Lastly, the \textit{Existing Images} category comprises non-copyrighted visuals collected from publicly available sources, used to expand the dataset with naturally occurring yet contextually unanswerable scenarios. To enable a more in-depth analysis of model performance, each image category was further divided into five subcategories of questions. This finer-grained structure allows us to examine whether models demonstrate particular strengths or weaknesses within specific types of reasoning challenges. Correspondingly, the visual questions were carefully curated to align with these subcategories, ensuring a consistent and systematic evaluation framework across all categories. It has been greatly discussed in Section \ref{sec:qtype_appendix}.


We categorized the \textbf{Surreal Images} into five subtypes based on the reasoning challenges they present: (1) \textit{Function Inquiry}, which questions the plausibility of an object’s use; (2) \textit{Component Inquiry}, focusing on missing or distorted essential parts; (3) \textit{Structural Stability Inquiry}, addressing physically unfeasible structures; (4) \textit{Material Compatibility Inquiry}, involving unrealistic material properties; and (5) \textit{Sensory Function Inquiry}, which challenges sensory expectations like heat or texture.

In a similar vein, the \textbf{Fusion of Objects and Animals} category was also divided into five subtypes to capture different dimensions of implausibility in hybrid entities. These include: (1) \textit{Anatomical Function Inquiry}, which examines the viability of altered physiological features; (2) \textit{Dietary Compatibility Inquiry}, exploring the logical consistency of feeding behaviors in mixed-species forms; (3) \textit{Mobility Inquiry}, addressing challenges in locomotion due to incompatible anatomical elements; (4) \textit{Communication Inquiry}, which questions the mechanisms of sound or signal production in hybrids; and (5) \textit{Adaptation Inquiry}, focusing on the feasibility of environmental integration or survival traits. This structured breakdown enables a deeper assessment of how models respond to biologically and mechanically incongruent scenarios.

To complement the surreal and fusion categories, we included the \textbf{Existing Figures} category featuring well-known fictional or mythological characters, paired with conceptually challenging questions. These were grouped into five subtypes: (1) \textit{Identity and Existence Paradoxes}, exploring contradictions in self-awareness or identity; (2) \textit{Time and Causality Loops}, involving paradoxes or alternate timelines; (3) \textit{Logic and Physics Violations}, breaking physical or narrative laws; (4) \textit{Reality and Fiction Blending}, mixing fictional logic with real-world constraints; and (5) \textit{Ethical and Philosophical Dilemmas}, raising questions of morality and agency. This category evaluates how well models handle abstract, high-level reasoning grounded in familiar yet paradoxical contexts. We include this category to examine whether models rely on memorized knowledge when presented with familiar characters commonly found online, enabling us to test their ability to distinguish between visual grounding and prior knowledge.

\begin{figure}[t]
    \centering
  \includegraphics[width=.85\columnwidth]{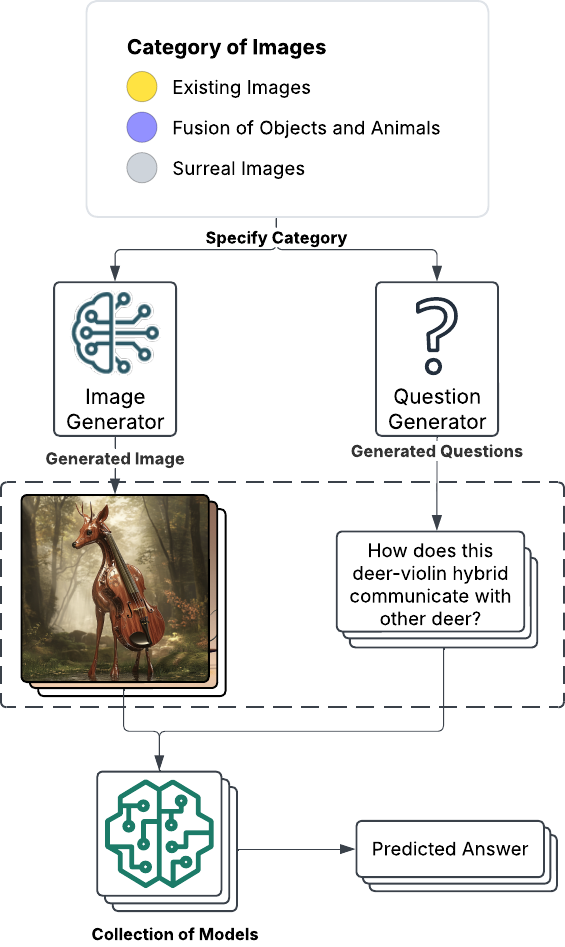}
  \caption{Overview of the pipeline for generating and evaluating questions on synthetic images. }
  \label{fig:flowchart}
\end{figure}


\subsection{Prompts and models for evaluation}
Prompt design plays a critical role in assessing whether models can recognize and appropriately handle unanswerable questions. To this end, we designed two standardized prompts demonstrated in \cref{fig:prompt}, where the model selects the most appropriate answer from four options or just answers by itself and provides a one-line justification. We also noted in the prompt that model predictions may align with the \textit{uncertain set}, indicating that the models may interpret certain questions as unanswerable. A similar approach was employed in the work of Bingbing \etal \cite{wen2024characterizing}, and the uncertain set used in our analysis was obtained from the work of Yuhong \etal  \cite{sun2024benchmarking}. 

We have investigated how large-scale models perform on our dataset. For our experiments, we evaluated LLaVA 7B \cite{liu2023visualinstructiontuning}, GPT-4o \cite{openai2024chatgpt}, GPT-4.1, and Gemini Flash 2.5 \cite{google2024gemini}. While these models have achieved strong performance on standard VQA benchmarks, applying them to our dataset revealed novel insights into their behavior and limitations, particularly in handling unanswerable visual questions.

\input{tablePrompt}

\subsection{Evaluation Metrics} 
Madhusudhan \etal \cite{madhusudhan2024llms} introduced the idea of confusion matrix and judging the models based on \textbf{True Positive(TP)}, \textbf{False Positive(FP)}, \textbf{True Negative(TN)} and \textbf{False Negative(FN)} illustrated in \cref{fig:confusion_matrix} and formulated the rate of abstention with both \textit{TN} and \textit{FN} . For our experiment we have defined \textbf{Abstention Rate (AR)}: 

\begin{equation}
\mathcal{AR} = \frac{X}{|\mathcal{D}|}, \quad X \in \{TN, FN\}
\label{eq:abstention_rate}
\end{equation}

Here,  $TN$ and $FN$ is the count of true negatives and false negatives and $|\mathcal{D}|$ is the total number of samples.

We designated the numeral `5' as the abstention marker, which is also demonstrated in \cref{fig:prompt}. This choice is grounded in empirical observations: during preliminary analysis, we noted that even when a model correctly identified a question as unanswerable in its reasoning, it often defaulted to producing a confident but incorrect answer. This behavior posed challenges in determining whether the model truly recognized unanswerability. However, we also observed that once models are guided to output the token `5' when they implicitly understood the unanswerability of a query. By standardizing `5' as the abstention signal, we are able to more reliably measure and compare abstention behavior across models, capturing their ability not only to solve but also to recognize the limits of their knowledge.

\section{Results}

\subsection{Abstention Behavior Across Models}

\cref{tab:performance} presents the abstention rates of four models--- LLaVA, GPT-4o, GPT-4.1, and Gemini 2.5 Flash, evaluated across two prompting settings: with and without answer options (\cref{fig:prompt}), and across three image categories. 
GPT-4o demonstrates the best overall abstention rates in both settings, particularly in the `Without Options' scenario, where it abstains from answering up to 93\% of the time on the `Fusion of Objects \& Animals' subset. This suggests a stronger capacity to recognize unanswerability, especially when not constrained by multiple-choice options. 

\begin{table}[ht]
\centering
\caption{Performance analysis on Abstention Behavior. \textcolor{green!50}{Green} and \textcolor{red!50}{red} demonstrates the best and the worst for each image category. }
\label{tab:performance}

\resizebox{0.49\textwidth}{!}{%
\begin{tabular}{C{1.3cm}|C{1cm}C{0.9cm}C{1cm}|C{1cm}C{0.9cm}C{1cm}}
\toprule
\multirow{2}{*}{\textbf{Model}}  & \multicolumn{3}{c}{\textbf{With Options}}  & \multicolumn{3}{c}{\textbf{Without Options}} \\
\cmidrule{2-7}
 &   \textbf{Existing} & \textbf{Fusion} & \textbf{Surreal} & \textbf{Existing} & \textbf{Fusion} & \textbf{Surreal} \\
\midrule

{LLaVA 7B}  & \cellcolor{red!20}0.0  & \cellcolor{red!20}0.03 & \cellcolor{red!20}0.04 & \cellcolor{green!20}0.954 & \cellcolor{green!20}0.972 & \cellcolor{green!20}0.98  \\
\midrule

{GPT-4o}  & \cellcolor{green!20}0.571 & \cellcolor{green!20}0.738 & \cellcolor{green!20}0.484 & 0.892 & 0.93 & 0.688 \\
\midrule

{GPT-4.1} &  0.144 & 0.336 & 0.258 & 0.792 & 0.814 & 0.501 \\
\midrule

{Gemini 2.5 Flash} &  0.152 & 0.29 & 0.292 & \cellcolor{red!20}0.61 & \cellcolor{red!20}0.696 & \cellcolor{red!20}0.466 \\
\bottomrule
\end{tabular}
} 
\small
\end{table}

GPT-4.1 also shows notable abstention behavior, especially in the `Without Options' setting, though its rates are consistently lower than GPT-4o. 
In contrast, Gemini Flash exhibits the lowest abstention rates across all categories, indicating a tendency to produce answers even in uncertain scenarios. These trends highlight GPT-4o’s more cautious and controlled response behavior compared to the more assertive, less abstention-prone outputs of Gemini Flash.

LLaVA exhibits a dual behavior, performing as either the best or the worst depending on the setting. While it shows an exceptionally high abstention rate of over 95\% across all categories in the absence of answer options, its performance deteriorates drastically when options are introduced—failing on almost all questions. This suggests that LLaVA is highly susceptible to being misled or biased when presented with multiple-choice options.

\subsection{Effect of `Option-Formatted' Questions on VLMs}

\cref{tab:performance} reveals a consistent and notable trend across all models: abstention rates are significantly higher in the without options setting compared to the with options condition. This suggests that models are more likely to correctly recognize unanswerable scenarios when they are not constrained by predefined choices. 
For example, GPT-4o shows a substantial increase in abstention from 0.571 to 0.892 on the `Existing' category and from 0.738 to 0.930 on the `Objects \& Animals' category when options are removed. A similar trend is observed for LLaVA, GPT-4.1 and Gemini Flash, though the magnitude varies.

\begin{figure}[h]
  \includegraphics[width=\columnwidth]{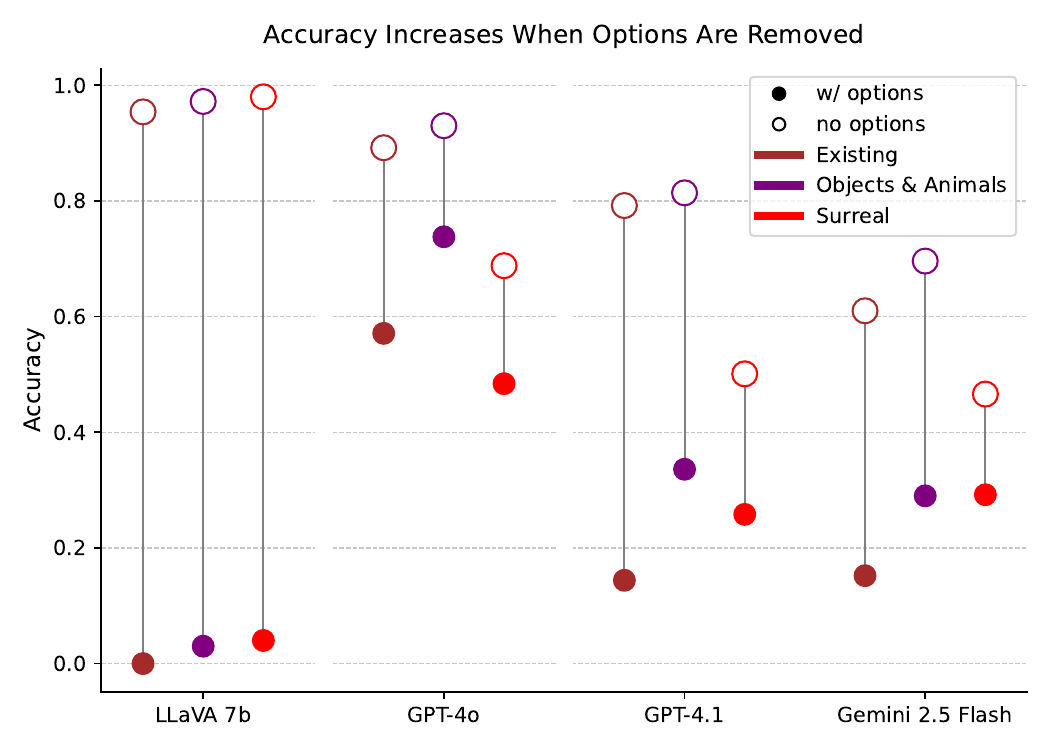}
  \caption{Abstention rates increase significantly when answer options are removed. Models demonstrate greater hesitation and uncertainty recognition in the no-options setting, particularly for visually ambiguous or surreal inputs. }
  \label{fig:acc_comparison}
\end{figure}

\cref{fig:acc_comparison} illustrates the shift in abstention behavior. LLaVA demonstrates the most significant deviation, transitioning abruptly from answering all questions to abstaining from answering altogether. GPT-4.1, which exhibits a 450\% increase in abstention accuracy on the `Existing' image category. This substantial change suggests that GPT-4.1 is more inclined to provide an answer when options are presented—especially for questions grounded in recognizable, real-world content. However, when deprived of options, the model transitions toward recognizing the question as unanswerable, particularly when it cannot extract sufficient information from the visual input alone. 


\subsection{VLMs Incorrectly Abstain on Answerable Questions}

\begin{figure}[t]
  \includegraphics[width=.9\columnwidth]{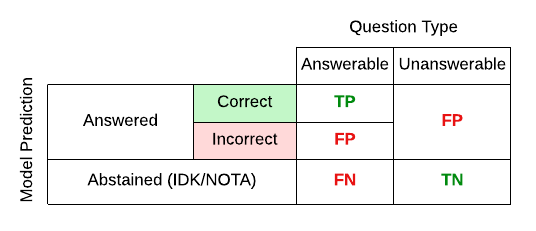}
  \caption{Confusion matrix that demonstrates True Positive, False Positive, True Negative and False Negative.}
  \label{fig:confusion_matrix}
\end{figure}

Our work focuses on how models respond when faced with inherently unanswerable questions--- a scenario that remains relatively underexplored due to the scarcity of such data. To investigate this, we use a carefully designed prompt, as illustrated in \cref{fig:prompt}. However, we also consider the opposite case: situations where the model abstains or responds incorrectly, despite the question being clearly answerable. 
For this analysis, we rely on the validation split of the \textbf{VQA v2.0} dataset \cite{goyal2017making}, using a subset of 1,000 questions. As shown in \cref{fig:confusion_matrix}, we define a false negative as a case where the model incorrectly identifies an answerable question as unanswerable. We used a prompt without answer options to encourage abstention on unanswerable questions. However, we also aim to examine whether this comes at the cost of performance on answerable ones.

\begin{figure}[ht]
    \centering
  \includegraphics[width=.9\columnwidth]{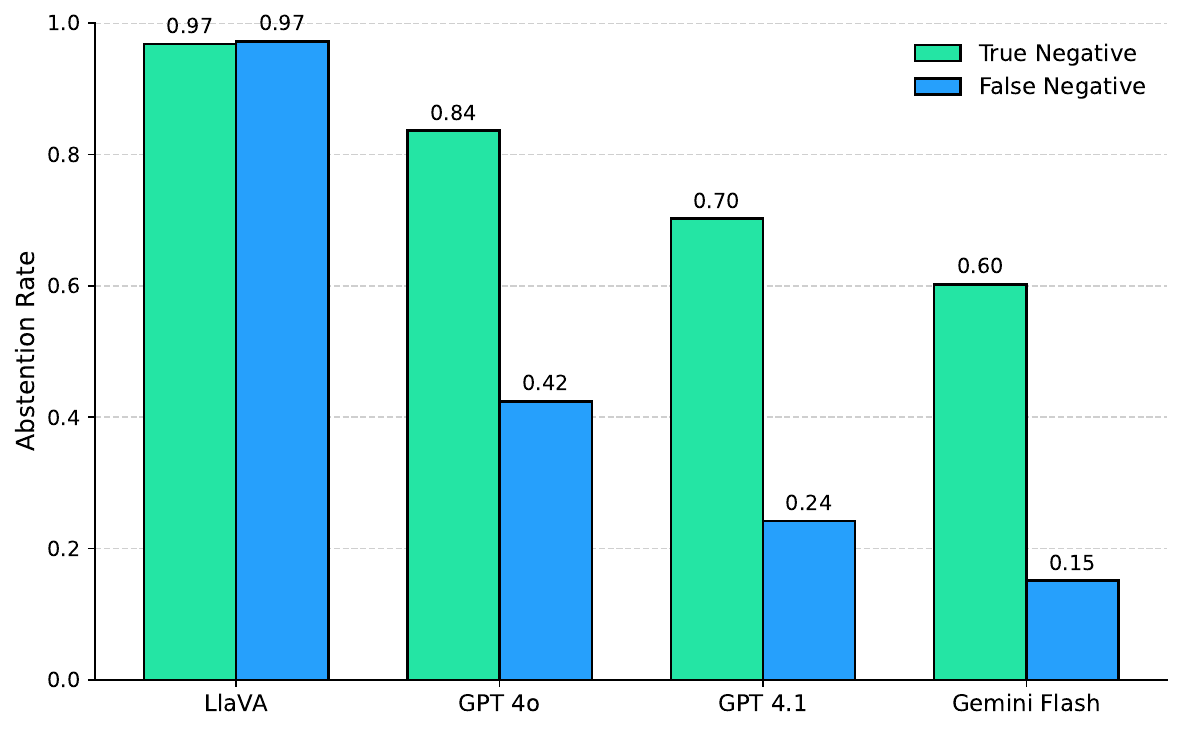}
  \caption{Abstention rates of different large language models (LLaVA, GPT-4o, GPT-4.1, and Gemini Flash) under True Negative and False Negative conditions}
  \label{fig:abstention_rate_bar}
\end{figure}

\begin{figure*}[t]
    \centering
  \includegraphics[width=.85\textwidth]{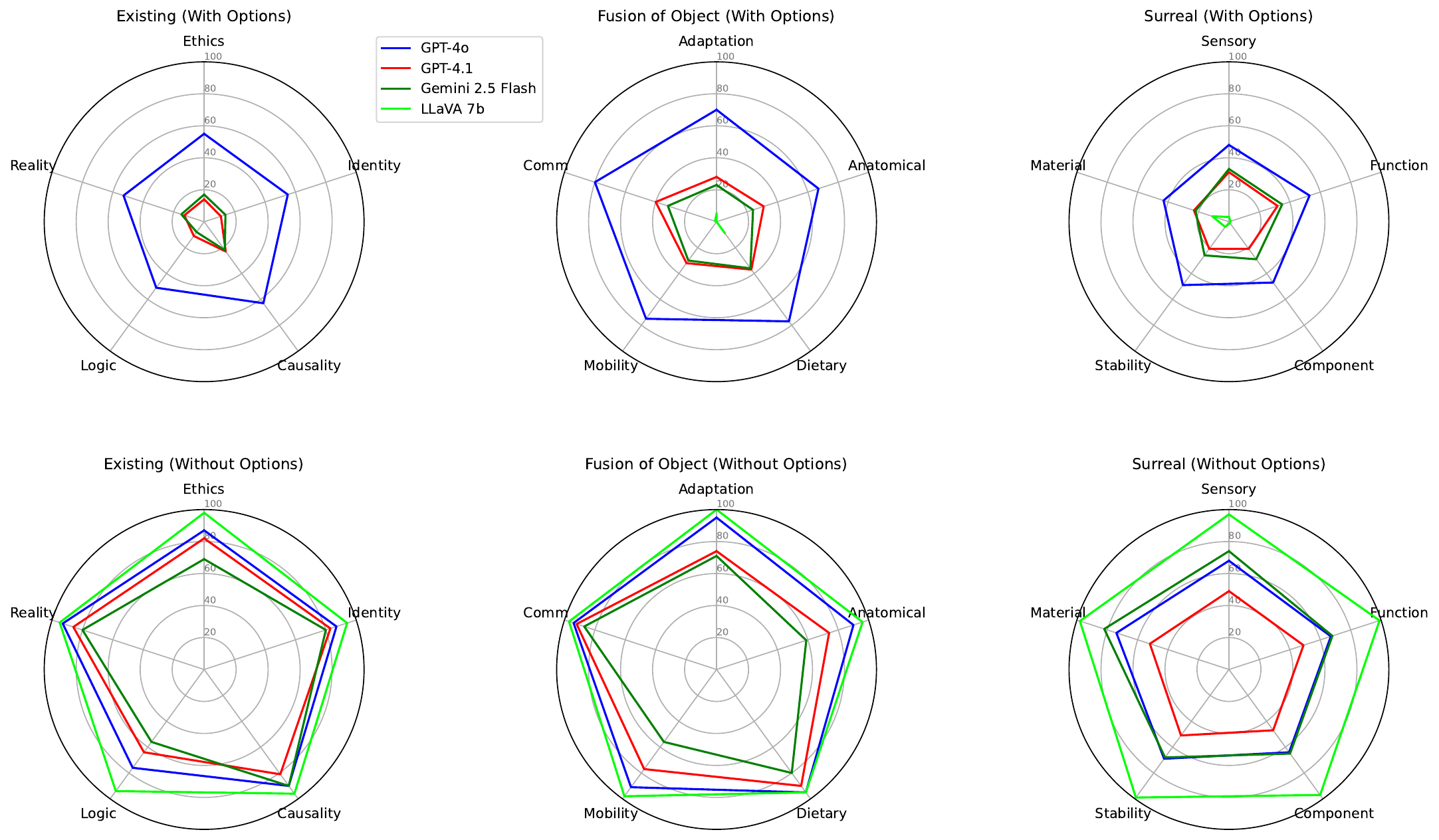}
  \caption{Performance comparison across different categories of illogical visual inputs, with and without answer options. Each radar chart demonstrates the abstention rate across different categories of questions. The top row represents performance with options, while the bottom row represents performance without options.}
  \label{fig:spider_man}
\end{figure*}

From \cref{fig:abstention_rate_bar}, we observe that \textbf{LLaVA} demonstrates the highest true negative rate (0.967), indicating a strong ability to correctly identify unanswerable questions. However, it also exhibits a relatively high false negative rate (0.972). This suggests that LLaVA fails to grasp the intended objective of determining answerability. The prompt explicitly instructs the model to mark a question as unanswerable only when appropriate; however, LLaVA surprisingly labels almost all questions as unanswerable. This behavior indicates a possible misunderstanding of the task objective conveyed by the prompt. Thus, relying solely on the abstention rate from unanswerable questions may not accurately reflect the model’s overall performance. 

\textbf{Gemini Flash}, while achieving the lowest true negative rate (0.60), also reports the lowest false negative rate (0.15), indicating a more risk-taking strategy that favors attempting answers even in borderline cases.

\subsection{Evaluating Abstention by Question Type}
We categorized the questions into distinct types and evaluated models both with and without answer options. Our goal was to investigate whether certain categories are more likely to confuse the models, particularly when distractor options are present. 
A detailed visualization of how each model performs across different question categories is shown in \cref{fig:spider_man}. When a model demonstrates a strong understanding of the image with the question, it tends to perform well across all categories; otherwise, it consistently fails to recognize unanswerable cases. A great example is GPT-4o as it performs well with whatever type of question it faces. The same cannot be said for LLaVA, GPT 4.1 or Gemini where they mostly get trapped or confused when presented with options, regardless of the question type they face. 

The most notable change is observed in Gemini, where the abstention rate increases significantly for \textbf{Dietary Compatibility Inquiry} and \textbf{Communication Inquiry} within the fusion of objects and animals category. This suggests that the model effectively identifies the absence of a viable solution given the image and the nature of the question.

\input{justificationTable}


\subsection{VLMs Justify the Unjustifiable}

While evaluating, we sought to go beyond simply recording the predicted answers. Rather than focusing solely on what the model answered, we aimed to understand why the model chose to answer at all, particularly when the appropriate response would have been to indicate the question was unanswerable. 
As illustrated in \cref{fig:prompt}, we collected the models' justifications for each response, even when the question lacked a valid ground truth. Our objective was to analyze and categorize these justifications in order to better understand the underlying reasoning strategies the models employed in these failure cases.

LLaVA was unable to give any proper justification as it failed to understand the given prompt. The other models provided justifications for unanswerable visual questions they mistakenly considered answerable, as illustrated in \cref{tab:justification_categories}. Our analysis revealed five distinct justification patterns. These patterns suggest that the models are not truly recognizing unanswerability but are instead conditioned to respond confidently due to training on large-scale datasets where every question has an answer. In this sense, the models are not hallucinating randomly, but overapplying their learned associations to force coherence where none exists.

\subsection{VLMs Inadequate Responses to Unanswerable Questions} 
We also evaluated other prominent vision-language models, including \textbf{PaliGemma} \cite{beyer2024paligemmaversatile3bvlm}, to assess their ability to handle unanswerable visual questions. Despite their impressive performance on standard benchmarks, our analysis revealed that these models consistently failed to interpret abstention-oriented prompts as intended. As shown in \cref{tab:justifications_LLaVA}, these are the justifications in sorted order provided by the PaliGemma model when presented with surreal images and unanswerable questions accompanied by answer options. Notably, the most frequent response that occurred in the majority of the 500 evaluated questions was \textbf{“Sorry, as a base VLM I am not trained to answer this question.”}.

Furthermore, as observed in \cref{tab:performance}, the LLaVA model fails to abstain from answering in 99\% of the cases with provided options. \cref{fig:compare_charts} further illustrates that the model predominantly selects option one, indicating a failure to comprehend the prompt and effectively identify unanswerable questions. This behavior suggests the presence of \textbf{recency bias} \cite{peysakhovich2023attentionsortingcombatsrecency}, where the model disproportionately favors the first available option, irrespective of its relevance.

This reveals a deeper limitation in distinguishing answerable from unanswerable inputs, especially beyond typical training distributions. It underscores the need for finer control over model confidence and abstention-aware training in future architectures.

\begin{table}[t]
\centering
\caption{Distribution of justifications with corresponding counts of PaliGemma.}
\begin{tabular}{p{6 cm} | c}
\toprule
\textbf{Justification} & \textbf{Count} \\
\midrule
Sorry, as a base VLM I am not trained to answer this question. & 245 \\
\midrule
The answer is uncertain. & 138 \\
\midrule
The answer is not relevant to the question. & 35 \\
\midrule
The answer is not available. & 27 \\
\midrule
The answer is not a question. & 16 \\
\midrule
The answer is not known. & 15 \\
\bottomrule
\end{tabular}

\label{tab:justifications_LLaVA}
\end{table}

\section{Conclusion} 

In this work, we investigated how VLMs respond to unanswerable visual questions, particularly in cases where the most appropriate behavior would be to abstain from answering. By designing a dataset called VisionTrap, we tested whether the VLMs used currently by a mass could recognize the boundaries of their visual and semantic understanding. We constructed our own diverse set of question categories to expose specific weaknesses in VQA models. The results suggest that models tend to hallucinate or produce confident answers even when the question is unanswerable, often defaulting to what appears to be the most plausible interpretation. Moreover, several VLMs are either not trained or inherently unable to effectively handle unanswerable questions. 
Building more challenging benchmarks is essential, as VQA remains an open problem. Although VLMs perform well on most datasets, our findings reveal a critical gap in their ability to recognize and admit the limits of their understanding.

{\small
\bibliographystyle{ieee_fullname}
\bibliography{references}
}

\section{Appendix}
\subsection{Question Types and Classification}
\label{sec:qtype_appendix}

Our dataset consists of three categories of images and five types of questions in each category.

\subsubsection{Surreal Images}
These images depict scenarios that could not possibly happen in the real world due to violations of how the physical world works or logical consistency. The scenes look strange, unrealistic, or dream-like. They may appear artistic or imaginative but are clearly not real.

\begin{figure*}[t]
  \includegraphics[width=\textwidth]{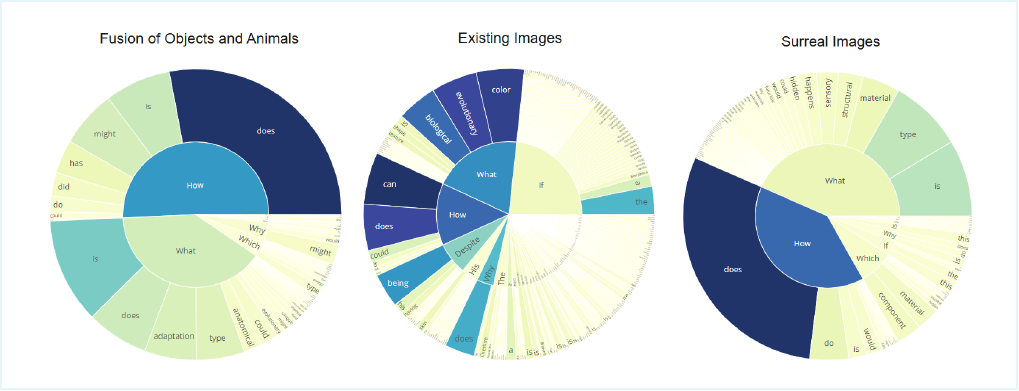}
  \caption{ Sunburst charts showing the distribution of question openers across three image categories: Fusion of Objects and Animals, Existing Images, and Surreal Images. Each chart visualizes the first and second words of questions, with segment size and color indicating frequency. Realistic images elicit more diverse linguistic structures, while surreal and fused-object images prompt more repetitive, interpretation-driven question forms.}
  \label{fig:sunburst}
\end{figure*}

Questions in this category can be divided into five subtypes:

\begin{enumerate}
    \item \textbf{Function Inquiry:} This category evaluates the plausibility of an object’s intended use or function within a given context. The questions focus on identifying the purpose or role of the object as depicted in the image, often assessing whether its practical utility aligns with the surrounding scenario. For example, if there is an image of soup served in a shoe, a function inquiry would be: \textit{How does the shoe hold the soup without spilling, given its original design as footwear?}
    
    \item \textbf{Component Inquiry:} This subtype focuses on missing or distorted essential components of an object. The questions assess how the design and presence of these components contribute to the object's practical usability in real-world scenarios.
    
    \item \textbf{Structural Stability Inquiry:} This category examines physically unfeasible structures, such as gravity-defying constructions or impossible geometries that contradict the principles of physical stability. The questions focus on how structural stability is obtained, despite the fact that such configurations would not be viable in real-world scenarios.
    
    \item \textbf{Material Compatibility Inquiry:} This category of questions inquires about unrealistic material properties, such as objects made from incompatible or contradictory substances. These questions challenge the model's ability to reason about material suitability within a given context.
    
    \item \textbf{Sensory Function Inquiry:} This subtype challenges the model's understanding of expected sensory experiences associated with objects. It inquires whether sensory attributes—such as heat, texture, or sound—are logically consistent with real-world experiences.
\end{enumerate}

\subsubsection{Fusion of Animals and Objects}
It consists of images where animals are unnaturally blended with inanimate objects, creating entities that resemble elements of the real world but do not exist in reality. The limitations of the models are tested through five types of questions:

\begin{figure*}[t]
  \centering
  \includegraphics[width=\textwidth]{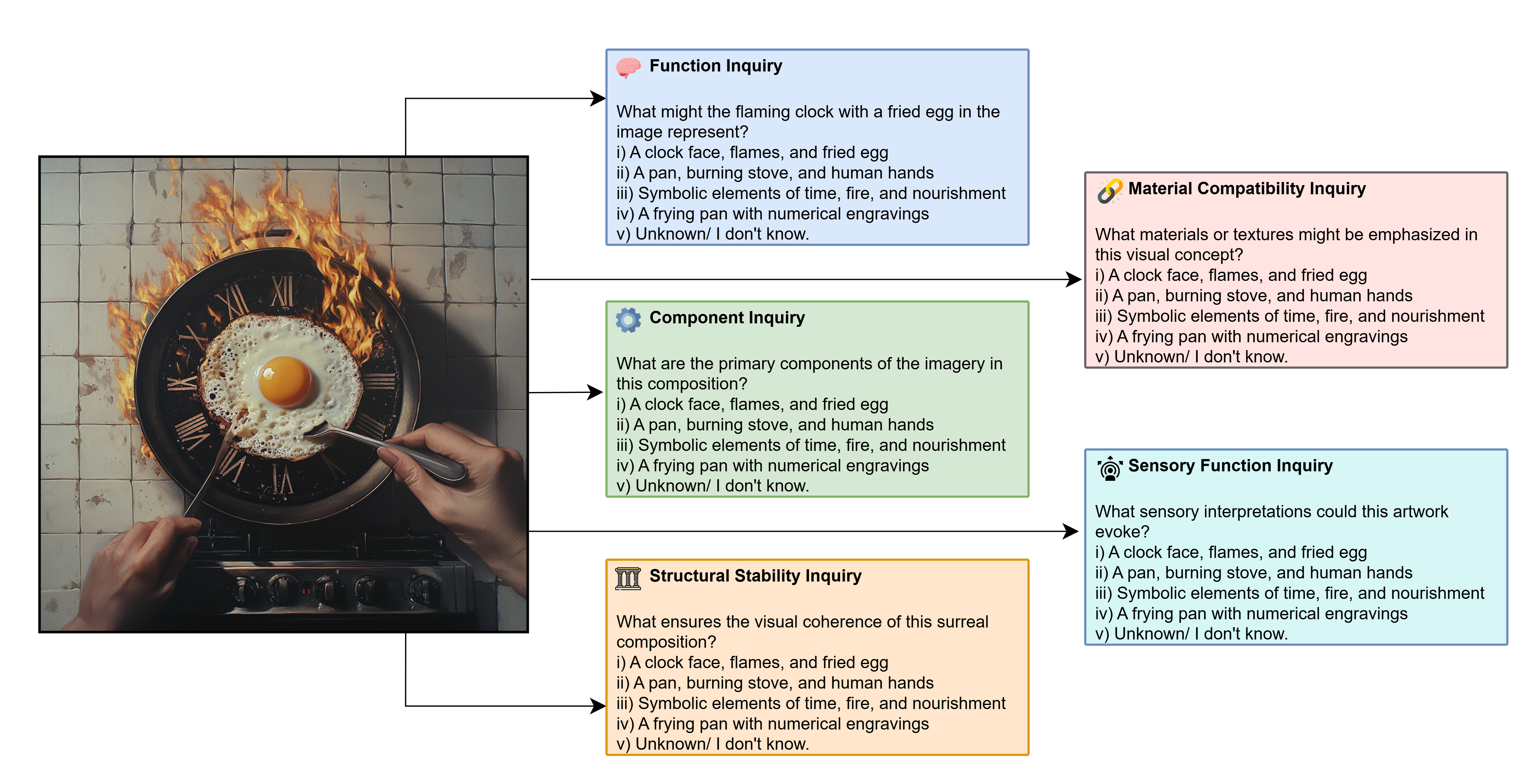}
  \caption{Example from the \textbf{Surreal Images} category with question types, where the image is intentionally unrealistic or dreamlike, often defying physical or logical laws}
  \label{fig:surreal_example}
\end{figure*}

\begin{enumerate}
    \item \textbf{Anatomical Function Inquiry:} This subtype examines the viability of altered physiological features in hybrid forms. Consider whether anatomical changes preserve or disrupt essential bodily functions of animals. Examines how changes in body structure still allow the animal to function properly in real life.

    \item \textbf{Dietary Compatibility Inquiry:} This explores the logical consistency of feeding behaviors in mixed species forms, assessing whether dietary habits from both sources can feasible co-exist. This involves assessing whether the digestive systems or metabolic processes of the animal can function properly given that it is fused with objects.

    \item \textbf{Mobility Inquiry:} This addresses the challenges of locomotion that arise from the combination of anatomically incompatible elements. Questions are related to how the animal moves or maintains balance in daily life, since they are not in their usual anatomical structure.

    \item \textbf{Communication Inquiry:} This subtype questions the mechanisms of sound or signal production in hybrids, investigating whether communication methods remain coherent or become biologically implausible. Assesses how animals interact in their daily lives.

    \item \textbf{Adaptation Inquiry:} This focuses on the feasibility of environmental integration or survival traits, evaluating whether the hybrid could realistically survive in any natural habitat. The questions are related to how the animals survive in their inherent ecosystem.
\end{enumerate}

\subsubsection{Existing Images}
 These images feature well-known fictional or mythological characters. The questions in this category are to test the high-level reasoning of models in paradoxical contexts. Although the images may be familiar to the models, the questions are unanswerable in a real-world context. The types of questions in this category are:

\begin{figure*}[t]
    \centering
  \includegraphics[width=.9\textwidth]{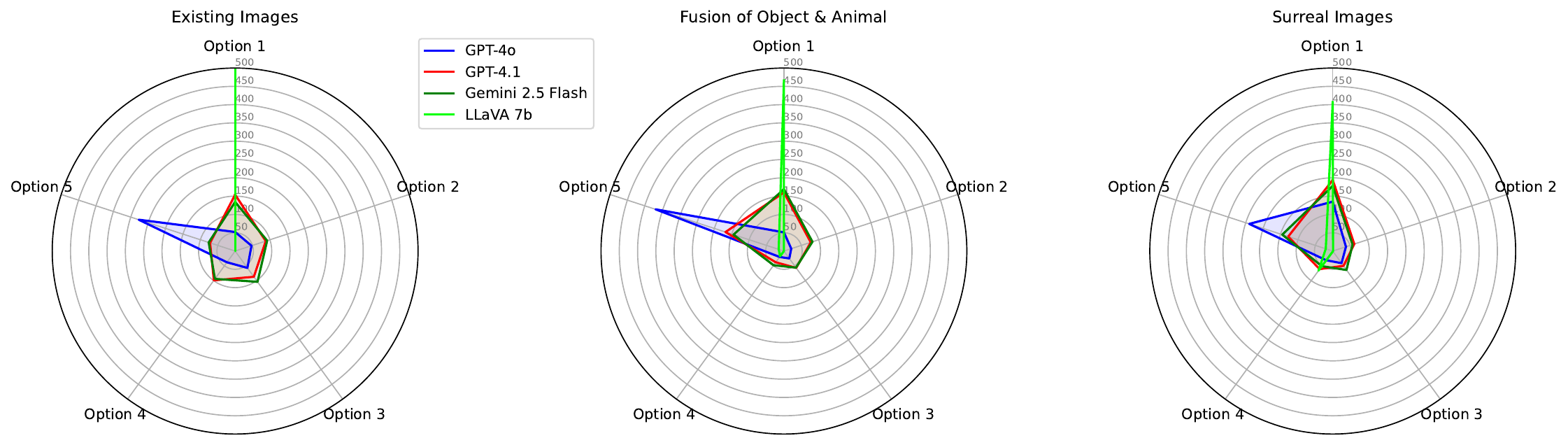}
  \caption{Answer distribution across multiple-choice options for four different model architectures over three question categories. Option 5 corresponds to the abstention choice—the correct response for all questions shown.}
  \label{fig:compare_charts}
\end{figure*}
 
\begin{enumerate}
    \item \textbf{Identity and Existence Paradoxes:}  
    This subtype explores contradictions in self-awareness or identity, such as a character questioning their own reality or continuity across versions. These questions create logical contradictions about the character’s identity, existence, or consciousness.

    \item \textbf{Time and Causality Loops:}  
    These involve paradoxes or alternate timelines, challenging the model to reason about events that disrupt chronological logic. Questions include scenarios involving time travel, causality paradoxes, or alternate versions of a character. For example, if Mickey Mouse meets his first black-and-white version, which one is more real?

    \item \textbf{Logic and Physics Violations:}  
    This category includes scenarios that violate established physical laws or logical consistency within the narrative. It covers questions involving time travel, paradoxes, teleportation inconsistencies, or alternate versions of a character that defy continuity or scientific principles.

    \item \textbf{Reality and Fiction Blending:}  
    This category involves scenarios where fictional logic is mixed with real-world constraints. These questions challenge the model to reconcile imaginative or fantasy-based rules---such as magical powers, futuristic technologies, or mythical settings---with realistic physical, ethical, or practical limitations found in the real world.

    \item \textbf{Ethical and Philosophical Dilemmas:}  
    This subtype presents scenarios that explore morality, personal agency, and difficult choices. These questions often place characters in situations that test their ethical beliefs, value systems, or sense of responsibility, raising deeper philosophical issues such as justice, free will, sacrifice, and the greater good.
\end{enumerate}

\subsection{Analysis of Question Types Across Image Categories}
To better understand the linguistic structure of questions posed in our dataset, we visualized their composition using sunburst charts demonstrated in Fig. \ref{fig:sunburst} based on the first two tokens of each question.

\subsubsection{Fusion of Objects and Animals}
In this category, questions are primarily initiated with How and What, which suggests a dominance of procedural and descriptive inquiries. The second words most commonly associated with \textit{How} include \textit{does, is,} and \textit{might}, indicating a strong presence of questions exploring behavior or hypothetical functionality of these fused entities. Questions like \textit{“How does it function?”} or \textit{“What type is this?”} are likely intended to probe the coherence or plausibility of object-animal hybrids.

\subsubsection{Existing Images}
This category exhibits the most lexical diversity among question starters. While \textit{What} remains the most frequent first token, we also observe a substantial number of questions beginning with \textit{If, How, Can,} and even atypical openers like \textit{Despite} or \textit{His}. The presence of content-heavy second words such as evolutionary, biological, and color suggests that these questions often relate to factual, scientific, or descriptive visual information. The broader distribution of tokens implies that annotators or users ask a wider variety of questions when the images are grounded in familiar, real-world contexts.

\subsubsection{Surreal Images}
Surreal images featuring illogical, fantastical, or physically impossible elements prompt a unique question distribution. While \textit{What} and \textit{How} still dominate, the second-word layer includes tokens like \textit{happens, sensory, hidden,} and \textit{structural}, reflecting interpretive or speculative inquiry. This aligns with the cognitive demand to rationalize implausible scenes. The frequency of \textit{“What is”} and \textit{“How does”} patterns implies that annotators attempt to extract meaning from abstract or conceptually challenging visuals.

\subsection{Answer Distribution Across Options and Abstention Behavior}
\cref{fig:compare_charts} illustrates the distribution of model-selected options across three categories: existing images, fusion of object and animal, and surreal images. Notably, \textbf{Option 5} corresponds to the abstention choice, which is the correct response for these unanswerable questions. As previously mentioned, LLaVA fails to abstain almost all of the questions and thus considers the option 1. GPT-4o shows a stronger preference for abstention in all three scenarios compared to GPT-4.0 and Gemini 2.5 Flash, particularly in cases involving object-animal fusion and surreal imagery. However, both GPT-4.0 and Gemini consistently exhibit a skew toward Option 1, suggesting a positional bias where the first available choice is disproportionately favored—regardless of its relevance or correctness. This behavior implies that models may rely on shallow heuristics or exhibit answer-order sensitivity, leading them to confidently choose a specific option even when abstention is more appropriate.

\subsection{Non-Response Behavior in Models}
In addition to explicit abstentions, we observed instances where models produced no response at all, neither an answer nor a justification. This behavior was most notable in GPT-4o, which, despite demonstrating strong overall performance, failed to produce any output for 39 questions in the with options setting. Such silent failures were not observed in the other models under the same condition. In contrast, Gemini 2.5 Flash exhibited a different pattern: in the without options setting, it failed to provide a justification in 520 out of 1,500 instances. 
However, this issue was largely absent when options were provided. This suggests that Gemini Flash is more likely to engage with the task when answer choices are explicitly presented, indicating a potential reliance on structured input to trigger response generation. 
Also, LLaVA consistently failed to provide justifications for most of its answers, suggesting a lack of understanding of the context or the expectation to justify its responses when required.

\subsection{Models for experiment}

We have investigated how large-scale models perform on our dataset. While these models have demonstrated strong results on widely used benchmarks, interesting insights on their capabilities were revealed being applied to our dataset.

\textbf{LLaVA} (Large Language and Vision Assistant) is an open-source vision-language model that integrates a pretrained language model (\eg., Vicuna \cite{zheng2023judging}) with visual encoders (typically CLIP-based) to enable multimodal understanding. It is trained using a combination of image-caption pairs and instruction-following data, allowing it to perform tasks such as visual question answering, image reasoning, and caption generation. Despite its strong performance on many benchmarks, LLaVA can be sensitive to prompt phrasing and may struggle with nuanced reasoning or ambiguous visual inputs.

\textbf{GPT-4o} is a multimodal model developed by OpenAI that achieves high performance across text, vision, and audio tasks while maintaining low latency and cost \cite{openai2023chatgpt}. Designed for efficient real-time applications, GPT-4o integrates the capabilities of GPT-4 with optimized inference and support for visual reasoning. 

\textbf{GPT-4.1} is an incremental update to OpenAI’s GPT-4 architecture, delivering improved reasoning, factual consistency, and task adaptability. Although OpenAI has not formally released detailed architectural specifications, public usage suggests enhancements in structured task handling and robustness to ambiguous queries.

\textbf{Gemini Flash 2.5}, released by Google DeepMind, is a lightweight variant of the Gemini 1.5 family optimized for fast and cost-efficient inference \cite{google2024gemini}. Despite its smaller size, it demonstrates competitive performance on many reasoning and coding benchmarks.



\subsection{Linguistic Markers of Confidence and Uncertainty}

To further understand the reasoning behavior of models when faced with unanswerable questions, we performed a linguistic analysis of the justifications generated by the models. Specifically, we examined the presence of \textbf{hedging words}, which indicate uncertainty (e.g., \textit{"might", "likely", "suggests"}) versus \textbf{confident words}, which signal assertiveness or factual claims (e.g., \textit{"is", "shows", "clearly"} ). Our analysis revealed a strong tendency toward confident language, with words such as "is" (2,033 occurrences), "are" (739), and "shows" (142) appearing far more frequently than hedging terms. In contrast, hedging phrases like "suggests" (222), "could" (166), and "likely" (123) were significantly less common. This imbalance indicates that the models often express high certainty, even when responding to logically unanswerable or ill-posed questions. Such linguistic overconfidence reflects a broader issue in current VQA systems: a lack of calibrated uncertainty, where models are incentivized to always provide an answer rather than acknowledge ambiguity or abstain. This highlights the need for future systems to incorporate uncertainty-aware training objectives and generate more cautious, appropriately hedged responses when confronted with uncertain or unanswerable inputs.

\end{document}

%% file: tablePrompt.tex
\begin{table}[t]
    \centering
    \caption{Comparison of prompt formats—with and without predefined option, along with response structure and uncertainty handling guidelines.}
    \begin{tabular}{>{\raggedright\arraybackslash}p{0.87\linewidth}}
        \toprule
       \rowcolor{lightgray} \textbf{Prompt with Options} \\
        \midrule
        \textbf{Answer Format:}
        \begin{itemize}
            \itemsep0em
            \item \textbf{Answer}: \textless just write the option number\textgreater
            \item \textbf{Justification}: \textless give one-liner justification\textgreater
        \end{itemize}
        \textbf{Note:} This is for research purposes only and your answer will not be used for any real-world decisions.

        \textbf{Question:} \{\textit{question}\}

        \textbf{Options:} \textit{option1}, \textit{option2}, \textit{option3}, \textit{option1}

        If you think your answer aligns with uncertain answers, then return \textbf{5}.\\
        Here are some examples of uncertain answers:
        \{\textit{uncertain-set}\} \\
        \midrule
        \rowcolor{lightgray}\textbf{Prompt without Options} \\
        \midrule
        \textbf{Answer Format:}
        \begin{itemize}
        \itemsep0em
            \item \textbf{Answer}: \textless just write the option number\textgreater
            \item \textbf{Justification}: \textless give one-liner justification\textgreater
        \end{itemize}
        \textbf{Note:} This is for research purposes only and your answer will not be used for any real-world decisions.

        \textbf{Question:} \{\textit{question}\}

        If you think your answer aligns with uncertain answers, then return \textbf{5}.\\
        Here are some examples of uncertain answers:
        \{\textit{uncertain-set}\} \\
        \midrule
        \textbf{\textit{uncertain-set}} = \{``The answer is unknown.'', ``The answer is uncertain.'', ``There is no definitive answer.'', ``It is not known.'', ``It is impossible to answer.''\} \\
        \bottomrule
    \end{tabular}
    \label{fig:prompt}
\end{table}

%% file: justificationTable.tex
\begin{table*}[t]
\centering
\small
\caption{Categories of model justifications on unanswerable questions, with examples showing reasoning patterns.\textcolor{red}{Red} emphasizes the portion for which the model decided to answer with justification.}
\label{tab:justification_categories}
\begin{tabular}{p{2.5cm}p{3.1cm}p{10.5cm}}
\toprule
\textbf{Category} & \textbf{Description} & \textbf{Example Justification of GPT 4o, 4.1 and Gemini} \\
\midrule

\textbf{Premise Denial or Logical Rebuttal} & Model rejects the question as illogical, implausible, or nonsensical. & 
``Goldfish \textcolor{red}{do not have dreams that can be decoded}'' \newline
``The premise of the question is \textcolor{red}{nonsensical}''\newline
``The scenario described is \textcolor{red}{not grounded in biological reality}'' \\
\midrule

\textbf{Visual Resemblance or Shape Matching} & Model identifies familiar objects based on visual similarity. & 
``The pencil-like structures \textcolor{red}{visually resemble pencils}'' \newline
``The object in the image \textcolor{red}{has a fish tail}, suggesting aquatic properties.'' \newline
``The form \textcolor{red}{mimics that of an eye}, implying perception or awareness.'' \\
\midrule

\textbf{Scene Composition or Spatial Context} & Model interprets spatial layout or interactions between objects. & 
``Books, furniture, and urban art are \textcolor{red}{arranged harmoniously} in the space.'' \newline
``The components are \textcolor{red}{placed to form a mechanical system}, implying function.'' \newline
``The clock and egg are \textcolor{red}{juxtaposed}, possibly representing a surreal moment in time.'' \\
\midrule

\textbf{Symbolic or Theoretical Interpretation} & Model interprets abstract or metaphorical meaning. & 
``The flaming clock with a fried egg likely \textcolor{red}{symbolizes surrealism} and the distortion of time.'' \newline
``Quantum foam composite \textcolor{red}{suggests theoretical possibilities} beyond current science.'' \newline
``The flames are a \textcolor{red}{manifestation of spiritual energy} in this depiction.'' \\
\midrule

\textbf{Pattern Recognition or Symmetry Analysis} & Model notices repeated patterns or symmetrical structures and assigns meaning. & 
``The arrangement of spoons forms a \textcolor{red}{symmetrical mandala-like pattern}.'' \newline
``The artwork shows metal shaped into an \textcolor{red}{intricate starburst design}.'' \newline
``Light bulbs connected by spokes \textcolor{red}{suggest a wheel-like formation}.'' \\
\bottomrule

\end{tabular}

\end{table*}